\documentclass[runningheads]{llncs}
\usepackage{framed}
\usepackage{mdframed}
\usepackage{xcolor}
\usepackage{svg}
\usepackage{booktabs}
\usepackage{enumitem}
\usepackage{algpseudocode}
\usepackage{multirow}
\usepackage{url}
\usepackage{framed}
\usepackage{graphicx}
\usepackage{caption}
\usepackage{subcaption}
\usepackage{balance}
\usepackage{makecell}
\usepackage[T1]{fontenc}
\usepackage{graphicx}
\usepackage{marvosym}
\usepackage[colorlinks,
            linkcolor=blue,
            anchorcolor=black,
            citecolor=blue]{hyperref}

\definecolor{shadecolor}{gray}{0.9}

\begin{document}
\title{AutoS$^2$earch: Unlocking the Reasoning Potential of Large Models for Web-based Source Search}
\titlerunning{Large Models for Web-based Source Search}
%
\author{Zhengqiu Zhu\inst{1} \and
Yatai Ji\inst{1} \and
Jiaheng Huang\inst{1} \and
Yong Zhao\inst{1} \and
Sihang Qiu\inst{1} \and 
Rusheng Ju\inst{1}
}
\authorrunning{Z. ZHU et al.}
%
\institute{National University of Defense Technology, Changsha, China \email{zhuzhengqiu@nudt.edu.cn,jiyatai\_1209@nudt.edu.cn,12254745966@qq.com,\ zhaoyong15@nudt.edu.cn,sihangq@acm.org,jrscy@sina.com}}
\maketitle              
\begin{abstract}

Web-based management systems have been widely used in risk control and industrial safety. However, effectively integrating source search capabilities into these systems, to enable decision-makers to locate and address the hazard (e.g., gas leak detection) remains a challenge. While prior efforts have explored using web crowdsourcing and AI algorithms for source search decision support, these approaches suffer from overheads in recruiting human participants and slow response times in time-sensitive situations.
To address this, we introduce AutoS$^2$earch, a novel framework leveraging large models for zero-shot source search in web applications. AutoS$^2$earch operates on a simplified visual environment projected through a web-based display, utilizing a chain-of-thought prompt designed to emulate human reasoning. The multi-modal large language model (MLLMs) dynamically converts visual observations into language descriptions, enabling the LLM to perform linguistic reasoning on four directional choices. Extensive experiments demonstrate that AutoS$^2$earch achieves performance nearly equivalent to human-AI collaborative source search while eliminating dependency on crowdsourced labor. Our work offers valuable insights in using web engineering to design such autonomous systems in other industrial applications.

\end{abstract}

\section{Introduction}

In today’s rapidly evolving digital landscape, the transformative power of web technologies has redefined not only how services are delivered but also how complex tasks are approached. Web-based systems have become increasingly prevalent in risk control across various domains. This widespread adoption is due their accessibility, scalability, and ability to remotely connect various types of users. For example, these systems are used for process safety management in industry~\cite{kannan2016web}, safety risk early warning in urban construction~\cite{ding2013development}, and safe monitoring of infrastructural systems~\cite{repetto2018web}. Within these web-based risk management systems, the source search problem presents a huge challenge. Source search refers to the task of identifying the origin of a risky event, such as a gas leak and the emission point of toxic substances. This source search capability is crucial for effective risk management and decision-making.

Traditional approaches to implementing source search capabilities into the web systems often rely on solely algorithmic solutions~\cite{ristic2016study}. These methods, while relatively straightforward to implement, often struggle to achieve acceptable performances due to algorithmic local optima and complex unknown environments~\cite{zhao2020searching}. More recently, web crowdsourcing has emerged as a promising alternative for tackling the source search problem by incorporating human efforts in these web systems on-the-fly~\cite{zhao2024user}. This approach outsources the task of addressing issues encountered during the source search process to human workers, leveraging their capabilities to enhance system performance.

These solutions often employ a human-AI collaborative way~\cite{zhao2023leveraging} where algorithms handle exploration-exploitation and report the encountered problems while human workers resolve complex decision-making bottlenecks to help the algorithms getting rid of local deadlocks~\cite{zhao2022crowd}. Although effective, this paradigm suffers from two inherent limitations: increased operational costs from continuous human intervention, and slow response times of human workers due to sequential decision-making. These challenges motivate our investigation into developing autonomous systems that preserve human-like reasoning capabilities while reducing dependency on massive crowdsourced labor.

Furthermore, recent advancements in large language models (LLMs)~\cite{chang2024survey} and multi-modal LLMs (MLLMs)~\cite{huang2023chatgpt} have unveiled promising avenues for addressing these challenges. One clear opportunity involves the seamless integration of visual understanding and linguistic reasoning for robust decision-making in search tasks. However, whether large models-assisted source search is really effective and efficient for improving the current source search algorithms~\cite{ji2022source} remains unknown. \textit{To address the research gap, we are particularly interested in answering the following two research questions in this work:}

\textbf{\textit{RQ1: }}How can source search capabilities be integrated into web-based systems to support decision-making in time-sensitive risk management scenarios? 

\textbf{\textit{RQ2: }}How can MLLMs and LLMs enhance the effectiveness and efficiency of existing source search algorithms? 


To answer the research questions, we propose a novel framework called Auto-\
S$^2$earch (\textbf{Auto}nomous \textbf{S}ource \textbf{Search}) and implement a prototype system that leverages advanced web technologies to simulate real-world conditions for zero-shot source search. Unlike traditional methods that rely on pre-defined heuristics or extensive human intervention, AutoS$^2$earch employs a carefully designed prompt that encapsulates human rationales, thereby guiding the MLLM to generate coherent and accurate scene descriptions from visual inputs about four directional choices. Based on these language-based descriptions, the LLM is enabled to determine the optimal directional choice through chain-of-thought (CoT) reasoning. Comprehensive empirical validation demonstrates that AutoS$^2$-\ 
earch achieves a success rate of 95–98\%, closely approaching the performance of human-AI collaborative search across 20 benchmark scenarios~\cite{zhao2023leveraging}. 

Our work indicates that the role of humans in future web crowdsourcing tasks may evolve from executors to validators or supervisors. Furthermore, incorporating explanations of LLM decisions into web-based system interfaces has the potential to help humans enhance task performance in risk control.

\section{Background and Motivation}
In this section, we review related works across three key areas and then outline the motivation behind this study.

\subsection{Web Crowdsourcing and Human-AI Collaboration Empowerment}
With the advancement of web technologies, crowdsourcing activities have increasingly migrated to web and mobile internet platforms, namely web crowdsourcing~\cite{doan2011crowdsourcing}. An exponential rise in its applications has witnessed, such as ride-hailing and software development.
To tackle complex web-based tasks, scientists at Microsoft introduced human-AI interaction guidelines to assist researchers and practitioners in designing studies utilizing AI technologies~\cite{amershi2019guidelines}. Following this, numerous studies have integrated human intelligence with AI methods to address challenges such as conversational agent learning for intent detection and text classification~\cite{yang2018leveraging,arous2021marta}. A recent study, for example, engaged online users from crowdsourcing platforms and implemented advanced computer vision techniques to generate city maps~\cite{qiu2019crowd}. Given the growing significance of AI-in-the-loop systems in human-intervened tasks, the concept and principles of human-AI decision-making within the context of web crowdsourcing were provided~\cite{green2019principles}. 


\subsection{Source Search and Crowd-powered Practices}
Source search is a critical problem for both nature and mankind~\cite{jing2021recent} focusing on determining the location of a source (of gas or signal) in the shortest possible time. Existing source search approaches can generally be classified into three categories: information-theoretic~\cite{jang2023improved}, biologically-inspired~\cite{al2021distributed}, and gradient-based methods~\cite{jiang2019source}. Among these, information-theoretic algorithms, especially those grounded in the Bayesian framework~\cite{ojeda2024robotic}, stand out for their distinct advantages. To further enhance the performance (i.e., success rate and efficiency) of a searching algorithm, multi-robot collaboration mechanisms~\cite{tang2020multirobot} have been designed and adopted. However, when source search takes place in complex environments, the search process always encounters fatal problems, resulting in wrong outcomes. Thus, researchers started to explore effective ways leveraging human intelligence to improve AI-based search algorithms through web platforms~\cite{zhao2024user}. However, this approach also entails substantial costs and imposes considerable burdens on human workers.

\subsection{Large Models for Scene Understanding and Reasoning}
MLLMs integrate multimodal encoders/decoders with traditional LLMs, enabling cross-modal understanding that overcomes text-only limitations. While these models demonstrate remarkable capabilities across diverse tasks including image-text understanding~\cite{liu2024visual}, video-text understanding~\cite{li2023videochat}, and even multimodal generation~\cite{peng2023kosmos}, their effectiveness in handling complex tasks remains constrained by predominant single-step reasoning approaches. To this end, CoT prompts are utilized to enhance problem-solving abilities by guiding LLMs through structured multi-step reasoning. Recent work explores CoT adaptations for multimodal problems, for instance, Shikra~\cite{chen2023shikra} pioneers CoT application in visual grounding tasks, while SoM~\cite{yang2023set} introduces structural image annotations like segmentation maps and spatial grids to provide spatial reasoning anchors. However, CoT has not been comprehensively explored for fine-grain reasoning in source search tasks.

\subsection{Motivation}
Building on the demonstrated scene understanding and reasoning capabilities of large models across various tasks, as well as addressing the limitations of human-AI collaborative source search, our work seeks to explore concrete methods for leveraging large models in zero-shot source search tasks within a top-down view of web-based search environments.

\section{System Design}
To answer \textbf{\textit{RQ1}}, we designed the AutoS$^2$earch framework based on web platforms. This involved migrating our previously developed crowd-powered source search prototype system~\cite{zhao2022crowd} from a desktop application to a web platform. The primary goals of this implementation were to achieve cross-platform accessibility, real-time interaction, and dynamic visualization.

(1) \textbf{Back-End Implementation:} we selected the lightweight and scalable Flask framework, and initialized the application using Flask and Socket.IO. The functions are handled by defining routes and Socket.IO events.


(2) \textbf{Real-Time Communication:} Socket.IO was used to support WebSocket and polling to ensure low-latency communication. Data is sent to the front-end using `socketio.emit', and on the front-end, events sent by the back-end are received using `socket.on'.

(3) \textbf{Map Drawing and Updating:} we first determined the map drawing logic (initializing the map and updating it based on changes), and then converted the map data into a format recognizable by the front-end.

(4) \textbf{Front-End Rendering:} we utilized HTML5's Canvas -- an ideal choice for dynamic map displays—to achieve efficient graphic rendering. We defined the Canvas element and implemented the drawing logic using JavaScript. To facilitate user interaction, we added control buttons on the interface and bound click events to them. Additionally, a click event is bound to the Canvas to send the user's click coordinates to the back-end.

\begin{figure}[htbp]
    \centering
    \includegraphics[width=\linewidth]{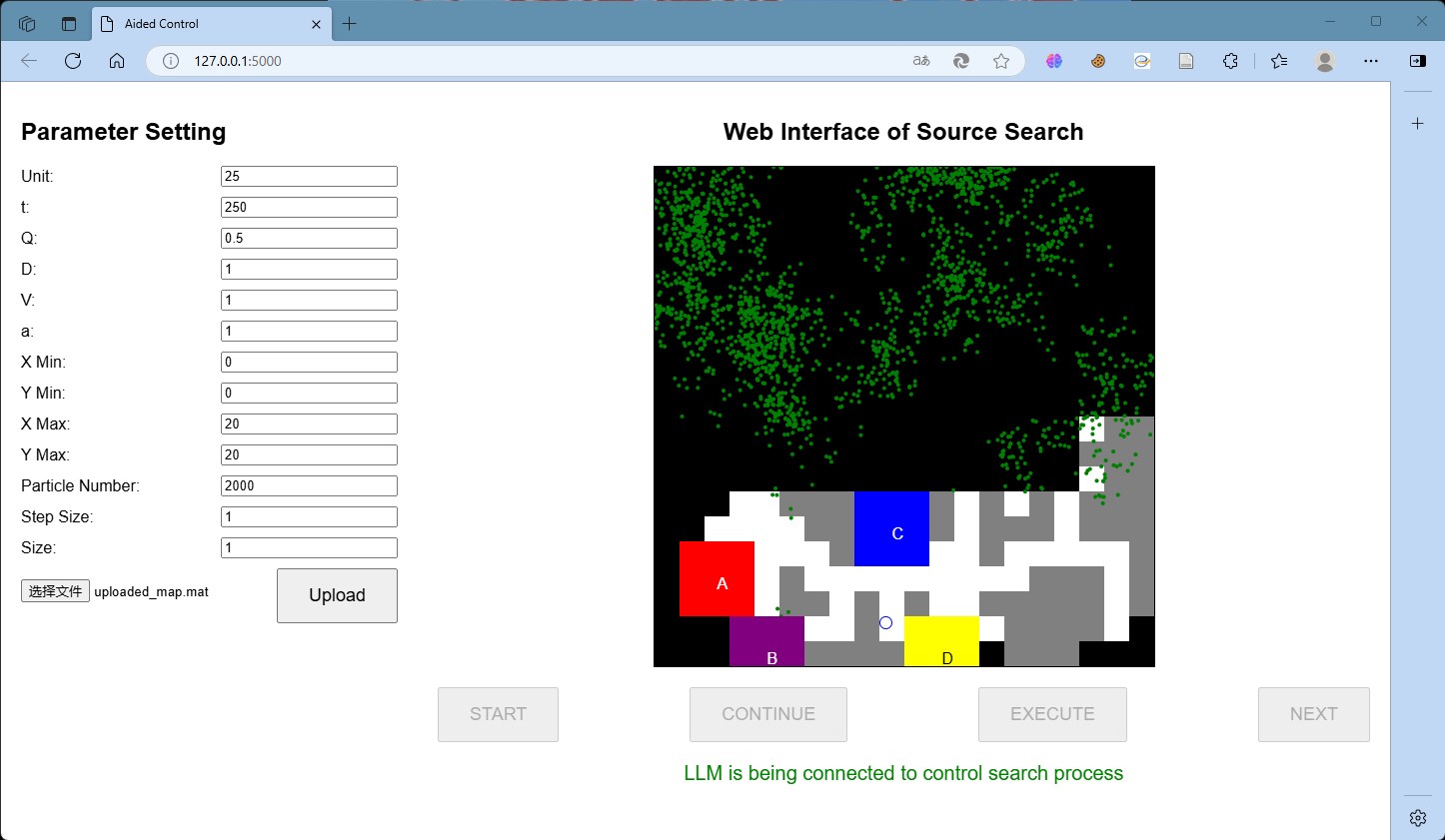}
    \caption{A screenshot of the web-based source search prototype system.}
    \label{fig:system}
\end{figure}

The user interface of this system, shown in Fig.~\ref{fig:system}, uses graphical elements to illustrate the source search task and the problem. It displays the robotic searcher, search environment, current state, posterior probability distribution of the source location, and four directional choices. When a problem is detected, the system automatically generates a task for large models, and then large models analyze the scene and reason step by step to plan a path for the robot. Once a deadlock is resolved, the system resumes automatic search, continuing until the source is found. This user interface allows decision-makers to observe the source search process in real-time. It provides the capability to: 1) enable decision-makers to interrupt the search process as needed, 2) facilitate crowdsourcing during the search, and 3) integrate large models for handling the detected problems.

\section{Method}
Previous studies have demonstrated the feasibility and effectiveness of human-AI collaborative source search in addressing fatal problems encountered in search problems (e.g., local optimum, dead end, infinite loop)~\cite{zhao2022crowd,zhao2023leveraging}. However, this approach comes with significant costs and imposes considerable burdens on human workers. Therefore, to answer \textbf{\textit{RQ2}}, we elaborate on the design of a large models-assisted source search method and explain how could it achieve human-like scene understanding and multi-step reasoning.

\subsection{Method Overview}
\begin{figure}[htbp]
    \centering
    \includegraphics[width=.95\linewidth]{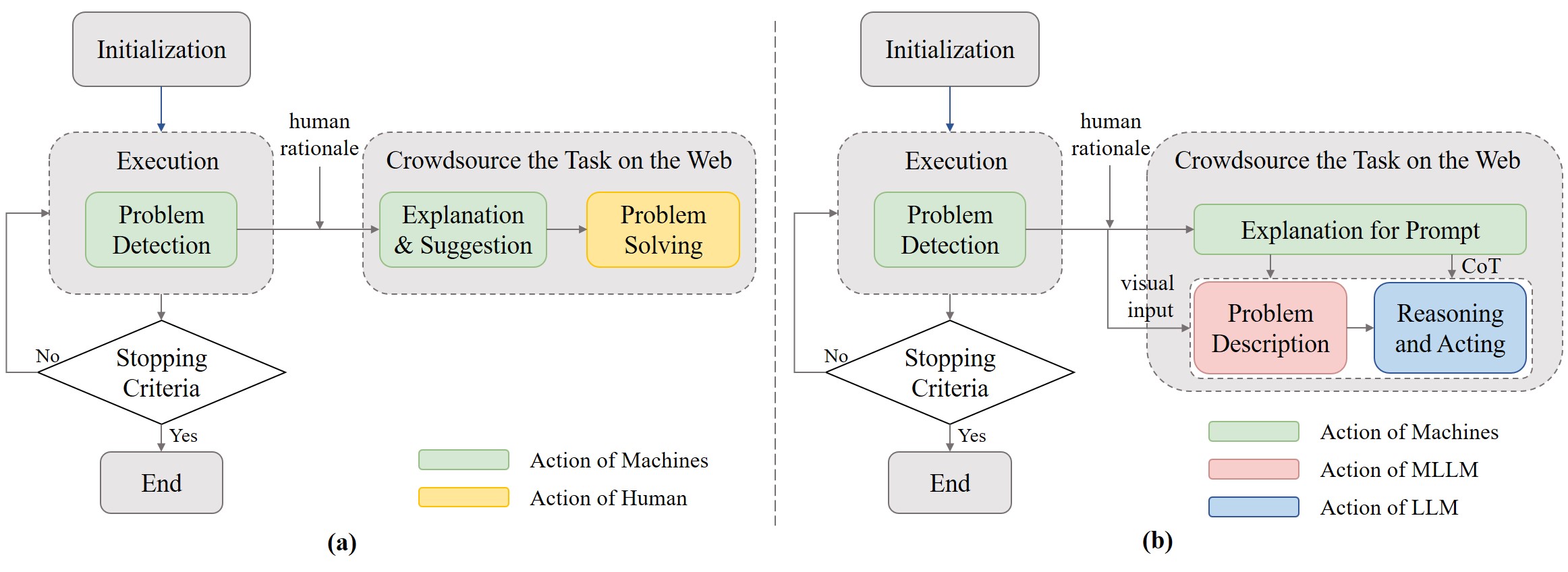}
    \caption{The schematic diagram of the AutoS$^2$earch method.}
    \label{fig:workflow}
\end{figure}

In this section, we show the design of AutoS$^2$earch, and introduce how MLLM and LLM can be used during the search process to improve the effectiveness and efficiency of search algorithms. There are various ways to achieve this goal in search. Here, we designed a straightforward workflow where the MLLM dynamically interprets visual data from a web-based display interface, converting it into detailed language descriptions for the LLM's CoT reasoning. Notably, no changes are made inside the search algorithm. The overview of the method is shown in as Fig.~\ref{fig:workflow}. Similar to the workflow of human-AI collaborative search (Fig.~\ref{fig:workflow}(a)), AutoS$^2$earch follows three main steps: initialization, execution, and end.\textit{ The main distinction lies in the execution phase, where AutoS$^2$earch incorporates four core components: machine-driven problem detection, machine-generated prompt explanations, problem description by the MLLM, and reasoning and acting by the LLM.} Except for the problem detection mechanism, the subsequent steps are totally different with those in our previous work~\cite{zhao2023leveraging}. In this work, we leverage human rationale to carefully design prompts, eliminating the need for human intervention after the method is initiated, as the problem-solving process is entirely handled by large models. Our approach helps reduce labor costs and accelerates problem-solving response time.  


\subsection{The Workflow Design of AutoS$^2$earch}

The prototype system was designed following the method outlined in Fig.~\ref{fig:workflow}(b), employing Infotaxis as the source searching algorithm. Infotaxis is one of the most popular cognitive search strategies, known for its effectiveness in solving source searching problems~\cite{ristic2016study}.

\textbf{(1) Problem Detection and Task Generation.} Through discussions with experts on the question, “\textit{what fatal problems could be happened during the search process}”, we have already identified common problems found in source search algorithms, such as local optima and deadlocks. Details can be found in \cite{zhao2023leveraging}. These issues can be detected in various ways, with one simple solution being a rule-based mechanism that automatically identifies the problematic search states and pauses the search process. A task is then generated and pushed to the web, seeking assistance from MLLMs and LLMs to facilitate effective problem-solving. A screenshot of the crowdsourcing task is shown in Fig.~\ref{fig:system}.

\begin{figure}[htbp]
    \centering
    \includegraphics[width=.85\linewidth]{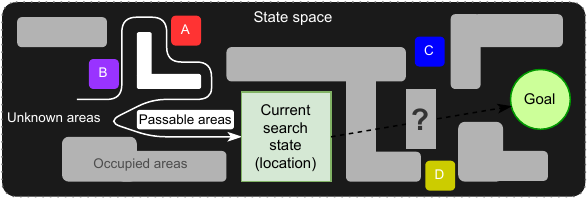}
    \caption{Elements needed to be presented while explaining the search algorithm and the problem.}
    \label{fig:explanation}
\end{figure}

\textbf{(2) Task Explanation for Prompt Design.} AI explanations hold great promise in the field of deep learning. However, explaining problems and search algorithms differs significantly from this domain. To help human workers or large models understand more intuitively, language and visuals are two highly effective means. While search algorithms are clear, the search process (e.g., search states) often remains hidden, making it hard to explain using language alone. Thus, we present key graphic elements to MLLMs or human workers by combining human rationale and AI algorithms, as shown in Fig.~\ref{fig:explanation}. These elements are identified through discussions with human experts and are specifically designed for providing explanations and suggestions. To understand the current situation, the robotic searcher's current position, occupied areas, passable areas, and unexplored areas should be prioritized for display. Additionally, to help large models better understand and make decisions, the task could provide a solution suggestion or some directional choices. The solution suggestion features a source estimation method that uses Bayesian inference and sequential Monte
Carlo methods to show the distribution of the posterior probability of the source location (see green particles in Fig.~\ref{fig:system}). Moreover, We also defined four regions—A, B, C, and D—around the robot's current position, representing the top-left, bottom-left, top-right, and bottom-right passable areas nearest to the unexplored areas. The potential goal and four directional choices are critical for large models to understand the problem and to reason better decisions.

\textbf{(3) Visual-Language Conversion by MLLM.}
When the task is explained through graphic elements, it comes to large models' turn to address the problem that a search algorithm cannot handle on its own. Using graphic elements as visual input, we specifically designed prompts for MLLMs. The main purpose of the prompt is to provide a hierarchical description of the situation around the four directional choices. The description is closely related to the search environment, current location of the searcher, and posterior probability distribution of the source location. The specific prompt for the MLLM is given.

\begin{center}
\begin{minipage}{\linewidth}
\begin{shaded}
\textit{\textbf{Prompt for MLLM}}

\textbf{Task Description:}  
Based on the graphic elements, identify the candidate target areas marked with letters in the image. Area A is marked in red; Area B in purple; Area C in blue; and Area D in yellow. Then provide a sequential description of each existing area, focusing on two main aspects for each:  
1. Distance to the dense green dot regions (classified as: Far, Medium, or Close)  
2. Density of surrounding unexplored black areas (classified as: High, Medium, or Low)  

\textbf{Output Format Example (assuming Area B does not exist):}  

Area A:  
Distance to dense green dot region: Far  
Density of surrounding unexplored areas: Low  

Area C:  
Distance to dense green dot region: Medium  
Density of surrounding unexplored areas: Medium  

Area D:  
Distance to dense green dot region: Close  
Density of surrounding unexplored areas: High

\textbf{Please describe the information for all existing areas in the image following this output format.}
\end{shaded}
\end{minipage}
\end{center}

\textbf{(4) Reasoning and Acting by LLM.} Based on the structured descriptions generated by the MLLM, we defined prioritized decision-making rules to enable the LLM to perform optimal strategy selection through chain-of-thought reasoning. These rules include: (1) Ignore non-existent areas (to prevent hallucinations of MLLMs); (2) Prioritize areas with the shortest distance to the dense green dot region; (3) Prioritize areas with the highest density of surrounding unexplored areas. The specific prompt for the LLM is given.

\begin{center}
\begin{minipage}{\linewidth}
\begin{shaded}
\textit{\textbf{CoT Prompt for LLM}}

\textbf{Task Description:}
Based on the structured descriptions provided by the MLLM, identify and output the single highest-priority area by following the rules below:

\textbf{Core Rules (in descending order of priority):}

1. Exclude non-existent regions.  

2. Prioritize the region closest to the dense green dot cluster.  

3. Prioritize the region with the highest density of surrounding unexplored black areas.

\textbf{Output Requirements:}

- Provide a single region letter (e.g., "C") as the result.

- Include a detailed reasoning process leading to the selection of the region.

\end{shaded}
\end{minipage}
\end{center}

\section{Experiments}\label{case}

In this section, we introduce experimental setup, baseline algorithms, and evaluation metrics. The project code can be found here\footnote[1]{https://gitee.com/parallelsimlab/autos2earch}.

\subsection{Experimental Setup}
The source search activities are performed by a virtual robot within a simulated 2D environment measuring $20m\times20m$. The search area is divided into a $20\times20$ grid of cells. Each cell has a probability $P_o$ of containing an obstacle, with $P_o$ set to 0.75 to introduce a relatively high difficulty (more obstacles). This higher complexity is chosen because simpler environments (with fewer obstacles) do not require external assistance. In this study, we did not consider the specific types or shapes of obstacles. If a cell contains an obstacle, it is considered completely obstructed, meaning the robot cannot enter or traverse it.

\subsection{Baseline Algorithms}
As detailed in the published work~\cite{zhao2023leveraging} on human-collaborative source search, the baselines adopted in this study naturally follow from that setup. Baseline 1 employs the Infotaxis algorithm directly, while Baseline 2 incorporates our proposed automatic problem detection method, navigating the robot to a random location to escape problematic scenarios. For consistency, we adopt an aided control interaction model of human-AI collaboration in this comparative analysis.
Furthermore, we introduce Baseline 3, where the robot navigates to a randomly chosen direction from four possible options (mentioned in Section 4.2) upon detecting a problem. It is worth noting that both Baseline 2 and Baseline 3 represent state-of-the-art improvements over traditional source search algorithms.

\subsection{Evaluation Metrics}
In this study, we evaluate the effectiveness and efficiency of the source search process and its outcomes. Effectiveness is measured by the success rate, defined as the robot successfully locating the source within 400 steps (where a step represents one iteration of updating search states). If the robot fails to find the source within 400 steps, regardless of whether large models are involved, the task is considered unsuccessful. Efficiency is assessed by the number of steps the robot takes to find the source, with failed attempts excluded from the calculation. Additionally, we measure the execution time of large models per task to see whether they hold an advantage over human workers in time-sensitive tasks.


\section{Results}
In this section, we present the results of (1) an illustrative run, (2) the comparison study, and (3) the ablation study. 

\begin{figure}[htbp]
    \centering
    \includegraphics[width=.7\linewidth]{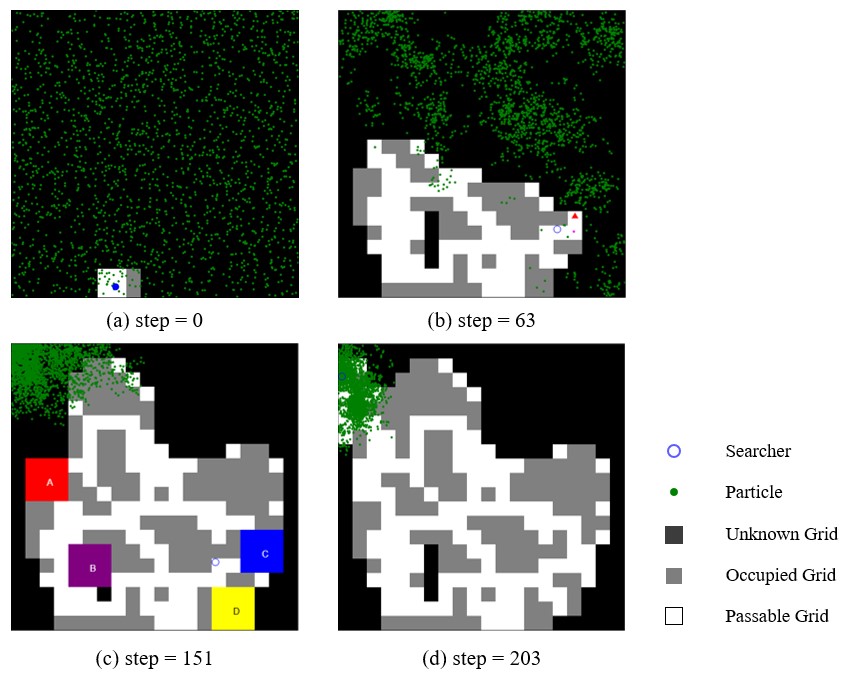}
    \caption{An illustrative run of the proposed framework at different time steps. (a) step=0; (b) step=63; (c) step=151; (d) step=203}
    \label{fig:Illus}
\end{figure}

\subsection{Illustrative Run} 

We conducted an experiment using one scenario from a set of 20 benchmark scenarios to illustrate a successful search process. The illustrative run of AutoS$^2$earch is shown in Fig.~\ref{fig:Illus}. The process includes the initiation of the search, the progression of the algorithm-driven search, the involvement of large models when a problem is detected, and ultimately resolving the issue to successfully locate the source. As we can see in Fig.~\ref{fig:Illus}(c), large models (both GPT-4o\footnote[2]{https://github.com/topics/gpt-4o}) are activated at search step=151. Based on the current visual inputs and the provided prompt for MLLMs, the language description of this scene is presented as follows. 

\begin{center}
\begin{minipage}{\linewidth}
\begin{shaded}
\textit{\textbf{Output of the MLLM}}

Region A:
Distance to dense area of small green dots: close;
Density of surrounding black exploration area: medium

Region B:
Distance to dense area of small green dots: medium;
Density of surrounding black exploration area: medium

Region C:
Distance to dense area of small green dots: medium;
Density of surrounding black exploration area: medium

Region D:
Distance to dense area of small green dots: far;
Density of surrounding black exploration area: low

\end{shaded}
\end{minipage}
\end{center}

Using the language descriptions of the four directional choices, the LLM initiates CoT reasoning, guided by the prompt for LLMs. The LLM's final choice is \textbf{A}, and its explanation for this selection is also provided. The decision-making process closely mirrors that of humans. 

To explore this further, we recruited three human raters to independently evaluate the LLM's reasoning process and compare the rationale behind the optimal choices made by the LLM and human intelligence. Our findings reveal that the LLM generally adheres strictly to a predefined priority order, whereas humans display an ability to flexibly adjust priorities based on the relative magnitudes of differences between the choices. For instance, when two regions are similarly distant from a dense cluster of green dots (indicating minimal differences in second-level priority), but the surrounding black area densities differ significantly (indicating greater differences in third-level priority), humans tend to prioritize the third-level criteria over the second. In contrast, the LLM struggle with such adaptive decision-making, even when additional rules are introduced.

\begin{center}
\begin{minipage}{\linewidth}
\begin{shaded}
\textit{\textbf{Reasoning Process of the LLM}}

Based on the description and priority rules, the analysis is conducted to identify the highest-priority region that meets the criteria:

1.First, check if any regions are missing:
Regions A, B, C, and D all exist.

2.Next, consider the closest distance to the dense area of green dots:
Region A is "close", regions B and C are "medium", and region D is "far"; thus, region A has the highest priority.

3.Then, consider the density of unexplored black areas surrounding each region:
Regions A, B, and C have "medium" density, Region D has "low" density.

\textbf{According to the priority rules, the region with the highest priority is A. Therefore, my selection is: A.}
\end{shaded}
\end{minipage}
\end{center}

\subsection{Comparative Study} 

In this subsection, we evaluated the effectiveness of large models-assisted source search over 20  benchmark scenarios (each scenario was run ten times) by measuring the effectiveness (success rate), the efficiency (the number of steps taken to find the source), the execution time of large models. The experimental results, averaged over Monte Carlo simulations across 20 scenarios, are presented in Table ~\ref{tab:results-1}. Clearly, large models-assisted source search has proved to be effective, achieving a success rate of 97\% in most cases. This represents an improvement of approximately 18.5\% over Baseline 1, 9\% over Baseline 2, and 7\% over Baseline 3, while being only 3\% lower than the success rate of human-AI collaborative search. Note that Baseline 2 and 3 are improvements based on the original algorithm (Baseline 1) since automatic problem detection and rule-based problem-solving strategies are used. Furthermore, we observe that the efficiency of AutoS$^2$earch (in terms of steps taken) is comparable to that of human-AI collaborative search, while the average execution time of large models is even shorter. For details on how human workers complete the crowdsourcing task, interested readers can refer to the previous work~\cite{zhao2022crowd}.

\begin{table}[!ht]
    \centering
    \caption{Results of the comparisons over various baselines.}
    \label{tab:results-1}
    \resizebox{0.9\textwidth}{!}{
    \begin{tabular}{llccc}
    \toprule
         \textbf{\emph{Methods}} & \textbf{Expertise} & \makecell[c]{\textbf{Effectiveness}\\(\% success rate)} & \makecell[c]{\textbf{Efficiency}\\(\# steps per task)} & \makecell[c]{\textbf{Human/MLLM+LLM execution time}\\(seconds per task)} \\
    \midrule
         \multirow{2}{*}{\emph{Human Aided}}
         & Expert   & 100  & 175.10 $\pm$ 67.67  & 33.58 $\pm$ 27.87\\
         & Non-expert  & 100  & 165.67 $\pm$ 80.60  & 29.01 $\pm$ 29.51\\
         \midrule
         \emph{Baseline 1} & -  & 78.5 & 154.04 $\pm$ 91.32  & - \\
         \midrule
         \emph{Baseline 2} & -  & 88  & 179.64 $\pm$ 96.45  & - \\
         \midrule
         \emph{Baseline 3} & -  & 90  & 179.76 $\pm$ 97.40  & - \\
         \midrule
         \emph{Ours} & -  & 97  & 170.97 $\pm$ 89.57  & 25.95 $\pm$ 38.20 \\
    \bottomrule
    \end{tabular}
    }
\end{table}

To further explore whether the impressive performance is solely due to GPT-4o's strong capabilities, we evaluated various combinations of MLLMs and LLMs from different companies. The results, presented in Table~\ref{tab:results-2}, reveal that our proposed framework is highly robust, consistently achieving success rates above 95\%. Notably, the Qwen model\footnote[3]{https://github.com/JMaiGC/ComfyUI-Qwen-VL-API} from the Chinese company Alibaba achieves the highest success rate at 98\%.

\vspace{-5mm}

\begin{table}[!ht]
    \centering
    \caption{Results of the comparisons over various large models.}
    \label{tab:results-2}
    \resizebox{0.9\textwidth}{!}{
    \begin{tabular}{llccc}
    \toprule
         \textbf{\emph{LLMs}} & \makecell[c]{\textbf{Effectiveness}\\(\% success rate)} & \makecell[c]{\textbf{Efficiency}\\(\# steps per task)} & \makecell[c]{\textbf{MLLM+LLM execution time}\\(seconds per task)} \\
    \midrule
         \emph{GLM-4v-plus + GLM-4-plus}  & 95 & 171.13 $\pm$ 92.64  & 26.39 $\pm$ 32.75 \\
         \midrule
         \emph{Qwen-VL-plus + Qwen-max}  & 98  & 172.85 $\pm$ 91.08  & 26.74 $\pm$ 36.39 \\
         \midrule
         \emph{GPT-4o + GPT-4o}  & 97  & 170.97 $\pm$ 89.57  & 25.95 $\pm$ 38.20 \\
    \bottomrule
    \end{tabular}
    }
\end{table}

\vspace{-5mm}

\subsection{Ablation Study}

We further ablation studies to validate the importance of main elements designed in our framework: the Chain-of-Thought prompt for the LLM and the size of directional choices A, B, C, and D (which determine the number of candidate cells for each option). We designated the model without CoT reasoning as Our-A and the model with reduced block sizes as Our-B. The average results across 20 scenarios are presented in Table \ref{tab:results-3}. As we can see, both the removal of CoT reasoning and the reduction in block sizes significantly decrease the success rate by approximately 6\% and 7\%, respectively. Notably, while removing CoT reasoning compromises the effectiveness performance, it does lead to improved efficiency and shorter execution time due to fewer reasoning steps.

\begin{table}[!ht]
    \centering
    \caption{Results of the ablation study.}
    \label{tab:results-3}
    \resizebox{0.8\textwidth}{!}{
    \begin{tabular}{llccc}
    \toprule
         \textbf{\emph{Methods}} & \makecell[c]{\textbf{Effectiveness}\\(\% success rate)} & \makecell[c]{\textbf{Efficiency}\\(\# steps per task)} & \makecell[c]{\textbf{MLLM+LLM execution time}\\(seconds per task)} \\
    \midrule
         \emph{Ours-A}  & 91 & 157.74 $\pm$ 85.33  & 23.82 $\pm$ 36.45 \\
         \midrule
         \emph{Ours-B}  & 90  & 170.28 $\pm$ 93.35  & 24.49 $\pm$ 34.19 \\
         \midrule
         \emph{Ours}  & 97  & 170.97 $\pm$ 89.57  & 25.95 $\pm$ 38.20 \\
    \bottomrule
    \end{tabular}
    }
\end{table}

\section{Discussions}

The source search results convey three main messages: (1) By incorporating carefully designed prompts that enable large language models with scene comprehension and multi-step reasoning capabilities, autonomous source search capabilities can be integrated into web-based systems to support decision-making in time-sensitive scenarios. (2) The large models-assisted method is effective and efficient for improving source search, approaching the performance of human-AI collaborative approaches while reducing execution time by approximately 25\%. (3) Whether in scene element presentation, problem detection mechanisms, or CoT prompt design, each component reflects human intelligence, highlighting that complex task solving fundamentally relies on human-AI hybrid intelligence.

\noindent\textbf{\textit{Drawbacks.}} Despite the strengths, this work has several limitations. (1) \textit{Environmental Complexity Gap:} The simplified $20 \times 20$ grid with static obstacles fail to capture real-world dynamics (e.g., moving obstructions, multi-source scenarios). The visual environment used here is insufficient to test whether large models truly possess robust scene understanding and multi-step reasoning capabilities in complex settings. (2) \textit{Limited Task Understanding:} While simple scene elements were designed to help the large model understand tasks, the lack of domain-specific knowledge makes it difficult for the model to balance exploration and exploitation during the search, sometimes leading to hallucinations by selecting irrelevant areas. (3) \textit{Underutilization of MLLM Potential}: In this work, MLLMs were mainly used to convert visual observations into textual descriptions, with large language models handling subsequent reasoning. This separation of visual understanding and language reasoning may limit the integrated capabilities MLLMs are designed to offer.

\noindent\textbf{\textit{Potential Avenues.}} To address these limitations, we propose to explore: (1) \textit{Dynamic Environment Adaptation:} Design LLM-empowered search agent and develop online prompt tuning mechanisms where LLMs could adjust decision rules according to the environment variations. (2) \textit{Visual Thinking Augmentation:} Integrate graph-based scene representations and reflection mechanisms to help MLLMs directly reason on the visual inputs without hallucinations. (3) \textit{Human-AI Value Alignment:} Implement human-in-the-loop feedback mechanisms in complex and high-risk scenarios and ensure alignment of decision objectives between humans and AI.

\noindent\textbf{\textit{Implications.}} The implications of AutoS$^2$earch extend far beyond the technical achievements in web-based autonomous systems. Its design reflects a broader trend in human-AI collaborative systems, where the goal is to harness the cognitive strengths of both entities in tandem.  Moreover, it may redefine the role of humans in web crowdsourcing systems—from task executors to validators of AI rationality in the future.

\section{Conclusions}

In this work, we present AutoS$^2$earch to address the issue of human dependency in web-based crowdsourcing systems for source search tasks. Through AutoS$^2$earch, we demonstrate that large models can effectively improve the performance of human-designed search algorithms in complex environments through visual-language translation and CoT reasoning. Our experimental validation shows AutoS$^2$earch achieves 95-98\% of human-AI collaborative source search algorithm effectiveness while eliminating labor costs and response time. \textit{This implies that modern large models can sufficiently replicate human scene reasoning for critical tasks like source search in complex environments.} As global industries increasingly lean on such systems for effective management, our work establishes a solid foundation for web engineering in other industrial applications.

\bibliographystyle{splncs04}
\bibliography{ref}

\begin{thebibliography}{10}
\providecommand{\url}[1]{\texttt{#1}}
\providecommand{\urlprefix}{URL }
\providecommand{\doi}[1]{https://doi.org/#1}

\bibitem{al2021distributed}
Al-Abri, S., Zhang, F.: A distributed active perception strategy for source seeking and level curve tracking. IEEE Transactions on Automatic Control  \textbf{67}(5),  2459--2465 (2021)

\bibitem{amershi2019guidelines}
Amershi, S., Weld, D., Vorvoreanu, M., Fourney, A., Nushi, B., Collisson, P., Suh, J., Iqbal, S., Bennett, P.N., Inkpen, K., et~al.: Guidelines for human-ai interaction. In: Proceedings of the 2019 chi conference on human factors in computing systems. pp. 1--13 (2019)

\bibitem{arous2021marta}
Arous, I., Dolamic, L., Yang, J., Bhardwaj, A., Cuccu, G., Cudr{\'e}-Mauroux, P.: Marta: Leveraging human rationales for explainable text classification. In: Proceedings of the AAAI conference on artificial intelligence. vol.~35, pp. 5868--5876 (2021)

\bibitem{chang2024survey}
Chang, Y., Wang, X., Wang, J., Wu, Y., Yang, L., Zhu, K., Chen, H., Yi, X., Wang, C., Wang, Y., et~al.: A survey on evaluation of large language models. ACM Transactions on Intelligent Systems and Technology  \textbf{15}(3),  1--45 (2024)

\bibitem{chen2023shikra}
Chen, K., Zhang, Z., Zeng, W., Zhang, R., Zhu, F., Zhao, R.: Shikra: Unleashing multimodal llm's referential dialogue magic (2023)

\bibitem{ding2013development}
Ding, L., Zhou, C.: Development of web-based system for safety risk early warning in urban metro construction. Automation in Construction  \textbf{34},  45--55 (2013)

\bibitem{doan2011crowdsourcing}
Doan, A., Ramakrishnan, R., Halevy, A.Y.: Crowdsourcing systems on the world-wide web. Communications of the ACM  \textbf{54}(4),  86--96 (2011)

\bibitem{green2019principles}
Green, B., Chen, Y.: The principles and limits of algorithm-in-the-loop decision making. Proceedings of the ACM on Human-Computer Interaction  \textbf{3}(CSCW),  1--24 (2019)

\bibitem{huang2023chatgpt}
Huang, H., Zheng, O., Wang, D., Yin, J., Wang, Z., Ding, S., Yin, H., Xu, C., Yang, R., Zheng, Q., et~al.: Chatgpt for shaping the future of dentistry: the potential of multi-modal large language model. International Journal of Oral Science  \textbf{15}(1), ~29 (2023)

\bibitem{jang2023improved}
Jang, H., Park, M., Oh, H.: Improved socialtaxis for information-theoretic source search using cooperative multiple agents in turbulent environments. Expert Systems with Applications  \textbf{225},  120033 (2023)

\bibitem{ji2022source}
Ji, Y., Zhao, Y., Chen, B., Zhu, Z., Liu, Y., Zhu, H., Qiu, S.: Source searching in unknown obstructed environments through source estimation, target determination, and path planning. Building and Environment  \textbf{221},  109266 (2022)

\bibitem{jiang2019source}
Jiang, X., Li, S., Luo, B., Meng, Q.: Source exploration for an under-actuated system: A control-theoretic paradigm. IEEE Transactions on Control Systems Technology  \textbf{28}(3),  1100--1107 (2019)

\bibitem{jing2021recent}
Jing, T., Meng, Q.H., Ishida, H.: Recent progress and trend of robot odor source localization. IEEJ Transactions on Electrical and Electronic Engineering  \textbf{16}(7),  938--953 (2021)

\bibitem{kannan2016web}
Kannan, P., Flechas, T., Mendez, E., Angarita, L., Chaudhari, P., Hong, Y., Mannan, M.S.: A web-based collection and analysis of process safety incidents. Journal of Loss Prevention in the Process Industries  \textbf{44},  171--192 (2016)

\bibitem{li2023videochat}
Li, K., He, Y., Wang, Y., Li, Y., Wang, W., Luo, P., Wang, Y., Wang, L., Qiao, Y.: Videochat: Chat-centric video understanding (2023)

\bibitem{liu2024visual}
Liu, H., Li, C., Wu, Q., Lee, Y.J.: Visual instruction tuning. Advances in neural information processing systems  \textbf{36} (2024)

\bibitem{ojeda2024robotic}
Ojeda, P., Monroy, J., Gonzalez-Jimenez, J.: Robotic gas source localization with probabilistic mapping and online dispersion simulation. IEEE Transactions on Robotics  (2024)

\bibitem{peng2023kosmos}
Peng, Z., Wang, W., Dong, L., Hao, Y., Huang, S., Ma, S., Wei, F.: Kosmos-2: Grounding multimodal large language models to the world (2023)

\bibitem{qiu2019crowd}
Qiu, S., Psyllidis, A., Bozzon, A., Houben, G.J.: Crowd-mapping urban objects from street-level imagery. In: The world wide web conference. pp. 1521--1531 (2019)

\bibitem{repetto2018web}
Repetto, M.P., Burlando, M., Solari, G., De~Gaetano, P., Pizzo, M., Tizzi, M.: A web-based gis platform for the safe management and risk assessment of complex structural and infrastructural systems exposed to wind. Advances in Engineering Software  \textbf{117},  29--45 (2018)

\bibitem{ristic2016study}
Ristic, B., Skvortsov, A., Gunatilaka, A.: A study of cognitive strategies for an autonomous search. Information Fusion  \textbf{28}, ~1--9 (2016)

\bibitem{tang2020multirobot}
Tang, H., Sun, W., Yu, H., Lin, A., Xue, M.: A multirobot target searching method based on bat algorithm in unknown environments. Expert Systems with Applications  \textbf{141},  112945 (2020)

\bibitem{yang2023set}
Yang, J., Zhang, H., Li, F., Zou, X., Li, C., Gao, J.: Set-of-mark prompting unleashes extraordinary visual grounding in gpt-4v (2023)

\bibitem{yang2018leveraging}
Yang, J., Drake, T., Damianou, A., Maarek, Y.: Leveraging crowdsourcing data for deep active learning an application: Learning intents in alexa. In: Proceedings of the 2018 World Wide Web Conference. pp. 23--32 (2018)

\bibitem{zhao2020searching}
Zhao, Y., Chen, B., Zhu, Z., Chen, F., Wang, Y., Ji, Y.: Searching the diffusive source in an unknown obstructed environment by cognitive strategies with forbidden areas. Building and environment  \textbf{186},  107349 (2020)

\bibitem{zhao2024user}
Zhao, Y., Ji, Y., Qiu, S., Zhu, Z., Ju, R.: A user interface design for collaborations between humans and intelligent vehicles. In: International Conference on Web Engineering. pp. 397--400. Springer (2024)

\bibitem{zhao2022crowd}
Zhao, Y., Zhu, Z., Chen, B., Qiu, S.: Crowd-powered source searching in complex environments. In: CCF Conference on Computer Supported Cooperative Work and Social Computing. pp. 201--215. Springer (2022)

\bibitem{zhao2023leveraging}
Zhao, Y., Zhu, Z., Chen, B., Qiu, S.: Leveraging human-ai collaboration in crowd-powered source search: a preliminary study. Journal of Social Computing  \textbf{4}(2),  95--111 (2023)

\end{thebibliography}

\end{document}